\documentclass[nohyperref]{article}

\usepackage{natbib}
\usepackage[top=1in, bottom=1in, left=1in, right=1in]{geometry}
\usepackage{etoc}
\usepackage{microtype}
\usepackage{graphicx}
\usepackage{xfrac}
\usepackage{subfigure}
\usepackage{booktabs}
\usepackage{multirow}
\usepackage{pifont}
\usepackage{wrapfig}
\usepackage{hyperref}
\usepackage{algorithm}
\usepackage{forloop}
\usepackage{tablefootnote}
\usepackage{algpseudocode}
\usepackage{enumitem}
\usepackage{amsmath}
\usepackage{amssymb}
\usepackage{mathtools}
\usepackage{amssymb}
\usepackage{pifont}
\newcommand{\cmark}{\ding{51}}
\newcommand{\xmark}{\ding{55}}
\newcommand{\tmark}{\ding{115}}
\usepackage{amsthm}
\usepackage[capitalize,noabbrev]{cleveref}
\usepackage{listings}
\usepackage{xcolor}
\definecolor{dkgreen}{rgb}{0,0.6,0}
\definecolor{dkred}{rgb}{0.8,0.0,0}
\definecolor{dkblue}{rgb}{0.0,0.0,0.9}
\definecolor{gray}{rgb}{0.5,0.5,0.5}
\definecolor{mauve}{rgb}{0.58,0,0.82}
\definecolor{lightgray}{HTML}{EEEEEE}
\definecolor{dkcyan}{HTML}{008b8b}
\definecolor{earthyellow}{rgb}{0.88, 0.66, 0.37}
\newtheorem{theorem}{Theorem}

\usepackage[font=small,labelfont=bf]{caption}
\usepackage{tikz}
\usepackage{wrapfig}
\usepackage{thm-restate}

\hypersetup{
    colorlinks=true,
    linkcolor=dkcyan,
    citecolor=dkcyan,
    filecolor=dkcyan,
    urlcolor=dkcyan
}
    
\everypar{\looseness=-1}
\setlist[itemize]{leftmargin=*}
\usepackage{amsmath,amsfonts,bm}

\def\eqref#1{equation~\ref{#1}}

\def\1{\bm{1}}

\def\rvtheta{{\boldsymbol{\theta}}}
\def\rvphi{{\boldsymbol{\phi}}}
\def\rvpsi{{\boldsymbol{\psi}}}

\def\rmB{{\mathbf{B}}}

\def\rmS{{\mathbf{S}}}

\def\vphi{{\bm{\phi}}}

\DeclareMathAlphabet{\mathsfit}{\encodingdefault}{\sfdefault}{m}{sl}
\SetMathAlphabet{\mathsfit}{bold}{\encodingdefault}{\sfdefault}{bx}{n}

\newcommand{\sigmoid}{\sigma}

\DeclareMathOperator*{\argmin}{arg\,min}

\newcommand{\Dtrain}{\mathcal{D}_\text{train}}
\newcommand{\batch}{\mathcal{B}}

\newcommand{\bmu}{\boldsymbol{\mu}}
\newcommand{\trace}{\text{tr}}
\newcommand{\bxi}{\boldsymbol{\xi}}

\newcommand{\loss}{\mathcal{L}}
\newcommand{\bx}{\mathbf{x}}
\newcommand{\bz}{\mathbf{z}}
\newcommand{\dkl}{D_\text{KL}}

\title{Multi-Rate VAE: Train Once, Get the Full Rate-Distortion Curve}

\author{%
Juhan Bae${}^{1, 2}$, 
Michael R. Zhang${}^{1, 2}$,  
Michael Ruan${}^{1}$,  
Eric Wang${}^{1}$,  \\
So Hasegawa${}^{3}$,  
Jimmy Ba${}^{1, 2}$ 
Roger Grosse${}^{1, 2, 4}$ \\ 
${}^{1}$University of Toronto, 
${}^{2}$Vector Institute, 
${}^{3}$Fujitsu Limited,
${}^{4}$Anthropic \\ 
\texttt{\{jbae, michael\}@cs.toronto.edu} 
}

\begin{document}
\etocdepthtag.toc{mtchapter}
\etocsettagdepth{mtchapter}{subsection}
\etocsettagdepth{mtappendix}{none}

\maketitle

\begin{abstract}
Variational autoencoders (VAEs) are powerful tools for learning latent representations of data used in a wide range of applications. In practice, VAEs usually require multiple training rounds to choose the amount of information the latent variable should retain. This trade-off between the reconstruction error (distortion) and the KL divergence (rate) is typically parameterized by a hyperparameter $\beta$. In this paper, we introduce Multi-Rate VAE (MR-VAE), a computationally efficient framework for learning optimal parameters corresponding to various $\beta$ in a single training run. The key idea is to explicitly formulate a response function that maps $\beta$ to the optimal parameters using hypernetworks. MR-VAEs construct a compact response hypernetwork where the pre-activations are conditionally gated based on $\beta$. We justify the proposed architecture by analyzing linear VAEs and showing that it can represent response functions exactly for linear VAEs. With the learned hypernetwork, MR-VAEs can construct the rate-distortion curve without additional training and can be deployed with significantly less hyperparameter tuning. Empirically, our approach is competitive and often exceeds the performance of multiple $\beta$-VAEs training with minimal computation and memory overheads. 
\end{abstract}
\section{Introduction}
\label{sec:introduction}

Deep latent variable models sample latent factors from a prior distribution and convert them to realistic data points using neural networks. Maximum likelihood estimation of the model parameters requires marginalization of the latent factors and is intractable to compute directly. \emph{Variational Autoencoders (VAEs)}~\citep{kingma2013auto,rezende2014stochastic} formulate a tractable lower bound for the log-likelihood and enable optimization of deep latent variable models by reparameterization of the \emph{Evidence Lower Bound (ELBO)}~\citep{jordan1999introduction}. VAEs have been applied in many different contexts, including text generation~\citep{bowman2015generating}, data augmentation generation~\citep{norouzi2020exemplar}, anomaly detection~\citep{an2015variational, park2022interpreting}, future frame prediction~\citep{castrejon2019improved}, image segmentation~\citep{kohl2018probabilistic}, and music generation~\citep{roberts2018hierarchical}.

In practice, VAEs are typically trained with the $\beta$-VAE objective~\citep{higgins2016beta} which balances the reconstruction error (\emph{distortion}) and the KL divergence term (\emph{rate}):
\begin{align}
    \loss_{\beta} (\rvphi, \rvtheta) = {\color{dkred} \underbrace{{\color{black} \mathbb{E}_{p_d (\mathbf{x})}[\mathbb{E}_{q_{\rvphi}(\bz | \bx)}[-\log p_{\rvtheta} (\bx| \bz)]]}}_{\text{Distortion}} } + \beta~{\color{dkgreen} \underbrace{ {\color{black} \mathbb{E}_{p_d (\mathbf{x})} [\dkl(q_{\rvphi}(\bz | \bx), p(\bz))]}}_{\text{Rate}}},
    \label{eq:beta_vae_objective}
\end{align}
where $p_{\rvtheta} (\bx| \bz)$ models the process that generates the data $\mathbf{x}$ given the latent variable $\mathbf{z}$ (the ``decoder'') and $q_{\rvphi}(\bz | \bx)$ is the variational distribution (the ``encoder''), parameterized by $\rvtheta \in \mathbb{R}^{m}$ and $\rvphi \in \mathbb{R}^p$, respectively. Here, $p(\mathbf{z})$ is a prior on the latent variables, $p_d(\mathbf{x})$ is the data distribution, and $\beta > 0$ is the weight on the KL term that trades off between rate and distortion. 

On one hand, models with low distortions can reconstruct data points with high quality but may generate unrealistic data points due to large discrepancies between variational distributions and priors~\citep{alemi2018fixing}. On the other hand, models with low rates have variational distributions close to the prior but may not have encoded enough useful information to reconstruct the data. Hence, the KL weight $\beta$ plays an important role in training VAEs and requires careful tuning for various applications~\citep{kohl2018probabilistic,castrejon2019improved,pong2019skew}. The KL weighting term also has a close connection to disentanglement quality~\citep{higgins2016beta,burgess2018understanding,nakagawa2021quantitative}, generalization ability~\citep{kumar2020implicit,bozkurt2021rate}, data compression~\citep{zhou2018variational,huang2020evaluating}, and posterior collapse~\citep{lucas2019don,dai2020usual,wang2022posterior}. 

By training multiple VAEs with different values of $\beta$, we can obtain different points on a rate-distortion curve (Pareto frontier) from information theory~\citep{alemi2018fixing}. Unfortunately, as rate-distortion curves depend on both the dataset and architecture, practitioners generally need to tune $\beta$ for each individual task. In this work, we introduce a modified VAE framework that does not require hyperparameter tuning on $\beta$ and can learn multiple VAEs with different rates in a single training run. Hence, we call our approach \emph{Multi-Rate VAE (MR-VAE)}.

We first formulate \emph{response functions} $\rvphi^{\star} (\beta)$ and $\rvtheta^{\star} (\beta)$~\citep{gibbons1992primer} which map the KL weight $\beta$ to the optimal encoder and decoder parameters trained with such $\beta$. Next, we explicitly construct response functions $\rvphi_{\rvpsi} (\beta)$ and $\rvtheta_{\rvpsi} (\beta)$ using hypernetworks~\citep{ha2016hypernetworks}, where $\rvpsi \in \mathbb{R}^h$ denotes hypernetwork parameters. Unlike the original VAE framework, which requires retraining the network to find optimal parameters for some particular $\beta$, response hypernetworks can directly learn this mapping and do not require further retraining.

\begin{figure}[t]
  \begin{minipage}[t]{0.39\linewidth}
    \centering
    \small
    \hspace{-1cm}
    \begin{tabular}{lccc}
    \hline
    \textbf{Methods} & \textbf{Parameters} & \begin{tabular}[c]{@{}c@{}}\textbf{Training}\\ \textbf{Time} (s) \end{tabular} & \begin{tabular}[c]{@{}c@{}}\textbf{RD}\\ \textbf{Curve}\end{tabular} \\ \hline
    \textbf{Standard VAE}              & $18.81 \times 10^{4}$ & 4,832  &  {\color{dkred} \xmark}\\
    \boldsymbol{$\beta$}\textbf{-VAE}               & $18.81 \times 10^{4}$ & 4,836  &  {\color{dkred} \xmark}\\
    \boldsymbol{$\beta$}\textbf{-VAEs} ($+8$ runs)     & $15.04 \times 10^{5}$ & 38,492 & {\color{earthyellow} \tmark} \\
    \boldsymbol{$\beta$}\textbf{-VAEs} ($+$ KL Annealing) & $15.04 \times 10^{5}$ & 38,640  &  {\color{earthyellow} \tmark} \\ \hline
    \textbf{MR-VAEs} (ours)            & $18.84 \times 10^{4}$ & 5,040   &  {\color{dkgreen} \cmark}  \\ 
    \hline
    \end{tabular}
    \par\vspace{0pt}
  \end{minipage}
  \begin{minipage}[r]{0.38\linewidth}
    \hspace{3.7cm}
    \vspace{-3cm}
    \includegraphics[width=\linewidth]{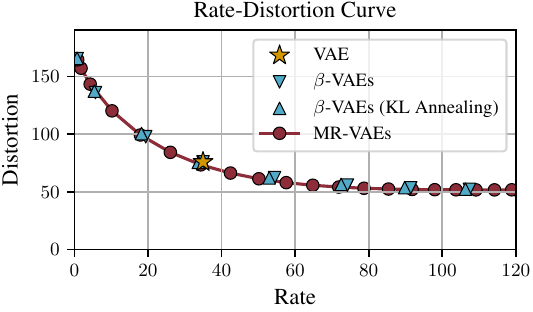}
  \end{minipage}
  \vspace{0.1cm}
  \caption{$\beta$-VAEs require multiple runs of training with different KL weights $\beta$ to visualize parts of the rate-distortion curve (Pareto frontier). Our proposed Multi-Rate VAEs (MR-VAEs) can learn the full continuous rate-distortion curve in a single run with a small memory and computational overhead.}
  \label{fig:mr_vae_fig1}
  \vspace{-1cm}
\end{figure}

While there is a lot of freedom in designing the response hypernetwork, we propose a hypernetwork parameterization that is memory and cost-efficient yet flexible enough to represent the optimal parameters over a wide range of KL weights. Specifically, in each layer of a VAE, our MR-VAE architecture applies an affine transformation to $\log \beta$ and uses it to scale the pre-activation. We justify the proposed architecture by analyzing linear VAEs and showing that the MR-VAE architecture can represent the response functions on this simplified model. We further propose a modified objective analogous to Self-Tuning Networks~\citep{mackay2019self,bae2020delta} to optimize response hypernetworks instead of the standard encoder and decoder parameters.

Empirically, we trained MR-VAEs to learn rate-distortion curves for image and text reconstruction tasks over a wide range of architectures. Across all tasks and architectures, MR-VAEs found competitive or even improved rate-distortion curves compared to the baseline method of retraining the network multiple times with different KL weights. We show a comparison between $\beta$-VAE (with and without KL annealing~\citep{bowman2015generating}) and MR-VAE with ResNet-based encoders and decoders~\citep{he2016deep} in Figure~\ref{fig:mr_vae_fig1}. MR-VAEs are capable of learning multiple optimal parameters corresponding to various KL weights in a single training run. Moreover, MR-VAEs do not require KL weight schedules and can be deployed without significant hyperparameter tuning. 

Our framework is general and can be extended to various existing VAE models. We demonstrate this flexibility by applying MR-VAEs to $\beta$-TCVAEs~\citep{chen2018isolating}, where we trade-off the reconstruction error and the total correlation instead of the reconstruction error and the rate. We show that MR-VAEs can be used to evaluate the disentanglement quality over a wide range of $\beta$ values without having to train $\beta$-TCVAEs multiple times.

\section{Background}
\label{sec:background}

\subsection{Variational Autoencoders}
\label{subsec:vae}

Variational Autoencoders jointly optimize encoder parameters $\rvphi$ and decoder parameters $\rvtheta$ to minimize the $\beta$-VAE objective defined in Eqn.~\ref{eq:beta_vae_objective}. While the standard ELBO sets the KL weight to 1, VAEs are commonly trained with varying KL weight $\beta$. As the KL weight $\beta$ approaches 0, the VAE objective resembles the deterministic autoencoder, putting more emphasis on minimizing the reconstruction error. On the other hand, when the KL weight is large, the objective prioritizes minimizing the KL divergence term and the variational distribution may collapse to the prior. 

It is possible to interpret VAEs from an information-theoretic framework~\citep{alemi2018fixing}, where the $\beta$-VAE objective is a special case of the information-bottleneck (IB) method~\citep{tishby2000information,alemi2016deep}. In this perspective, the decoder $p_{\rvtheta}(\mathbf{x}|\mathbf{z})$ serves as a lower bound for the mutual information $ I_{q}(\mathbf{x} ; \mathbf{z})$ and the rate upper bounds the mutual information. The relationship can be summarized as follows:
\begin{align}
    H - D \leq I_{q}(\mathbf{x} ; \mathbf{z}) \leq R,
\end{align}
where $D$ is the distortion, $H$ is the data entropy that measures the complexity of the dataset, and $R$ is the rate. Note that we adopt terminologies ``rate'' and ``distortion'' from rate-distortion theory~\citep{thomas2006elements,alemi2018fixing,bozkurt2021rate} which aims to minimize the rate under some constraint on distortion. The $\beta$-VAE objective in Eqn.~\ref{eq:beta_vae_objective} can be seen as a Lagrangian relaxation of the rate-distortion objective:
\begin{align}
    \loss_{\beta} (\rvphi, \rvtheta) = D + \beta R
\end{align}
In this view, training multiple VAEs with different values of $\beta$ corresponds to different points on the rate-distortion curve and distinguish models that do not utilize latent variables and models that make large use of latent variables~\citep{phuong2018mutual}. We refer readers to \citet{alemi2018fixing} for a more detailed discussion of the information-theoretic framework.

\subsection{Linear VAEs}
\label{subsec:linear_vae}

The linear VAE is a simple model where both the encoder and decoder are constrained to be affine transformations. More formally, we consider the following problem from~\citet{lucas2019don}:
\begin{equation}
  \begin{gathered}
    p_{\rvtheta}(\bx | \bz) = \mathcal{N}(\mathbf{D} \bz + \bmu, \sigma^2 \mathbf{I})\\
    q_{\rvphi}(\bz | \bx) = \mathcal{N}(\mathbf{E} (\bx - \bmu), \mathbf{C}),
  \end{gathered}
   \label{eq:linear_vae_setup}
\end{equation}
where $\mathbf{C}$ is a diagonal covariance matrix that is shared across all data points, $\mathbf{E}$ is the encoder weight, and $\mathbf{D}$ is the decoder weight. 

While the analytic solutions for deep latent models typically do not exist, the linear VAE has analytic solutions for optimal parameters and allows us to reason about various phenomena in training VAEs. For example,~\citet{dai2018connections} show the relationship of linear VAE, probabilistic PCA (pPCA)~\citep{tipping1999probabilistic}, and robust PCA~\citep{candes2011robust,chandrasekaran2011rank} and analyze the local minima smoothing effects of VAEs. Similarly,~\citet{lucas2019don} and \citet{wang2022posterior} use linear VAEs to investigate the cause of the posterior collapse~\citep{razavi2019preventing}. 

\subsection{Response Functions}
\label{subsec:response_function}

Response (rational reaction) functions~\citep{gibbons1992primer, lorraine2018stochastic} in neural networks map a set of hyperparameters to the optimal parameters trained with such hyperparameters. In the case of $\beta$-VAEs, response functions map the KL weight $\beta$ to optimal encoder and decoder parameters that minimize the $\beta$-VAE objective:
\begin{equation}
  \begin{gathered}
      \rvphi^\star (\beta), \rvtheta^\star (\beta) = \argmin_{\rvphi \in \mathbb{R}^p, \rvtheta \in \mathbb{R}^m} \loss_{\beta} (\rvphi, \rvtheta).
  \end{gathered}
\end{equation}
Approximation of response functions has various applications in machine learning, including hyperparameter optimization~\citep{lorraine2018stochastic}, game theory~\citep{fiez2019convergence,wang2019solving}, and influence estimation~\citep{bae2022if}.

\section{Methods}
\label{sec:methods}

In this section, we introduce Multi-Rate VAEs (MR-VAEs), an approach for directly modeling VAE response functions with hypernetworks. We first formulate memory and cost-efficient response hypernetworks $\rvphi_{\rvpsi} (\beta)$ and $\rvtheta_{\rvpsi} (\beta)$ and justify the proposed parameterization by showing that they recover exact response functions for linear VAEs. Then, we propose a modified training objective for MR-VAEs and formally describe an algorithm for learning the full rate-distortion curve in a single training run.

\subsection{Response Hypernetworks for VAEs}
\label{subsec:response_hypernetwork}

\begin{figure*}[t]
    \small
    \centering
    \vspace{-0.4cm}
    \includegraphics[width=0.9\linewidth]{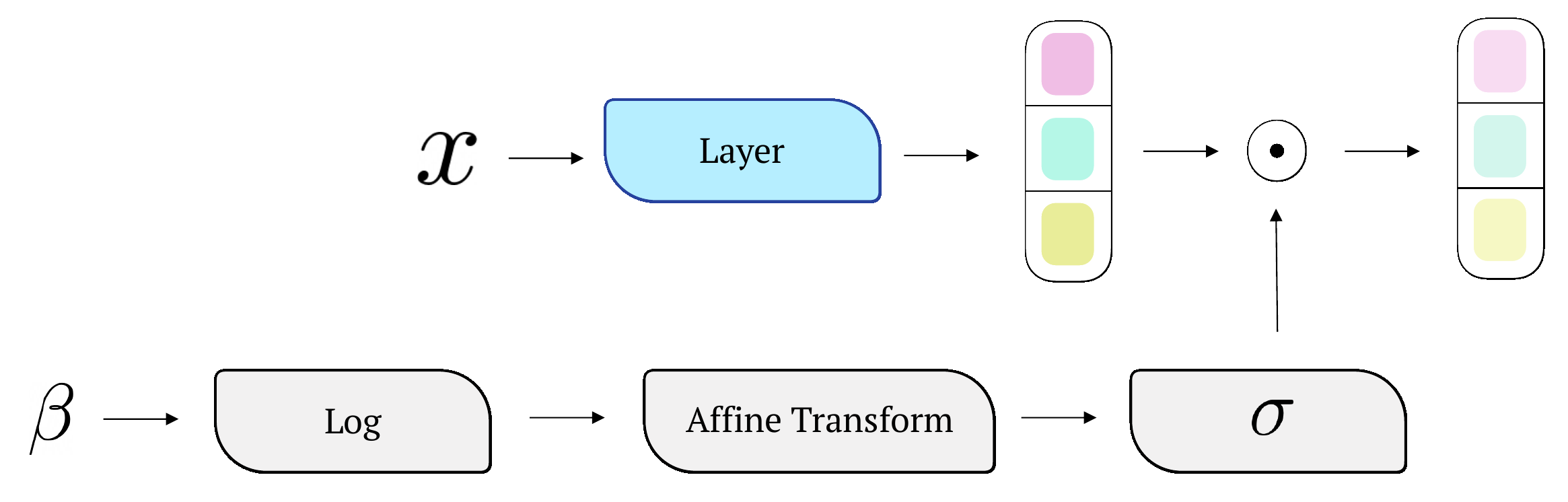}
    \caption{The MR-VAE architecture scales the pre-activations for each layer of the base VAE. This scaling term is generated with an affine transformation on the log KL weight followed by an activation function.}
    \label{fig:mr_vae_diagram}
    \vspace{-1.1cm}
\end{figure*}

\begin{wrapfigure}[14]{R}{0.4\textwidth}
    \centering
    \vspace{-0.9cm}
    \includegraphics[width=0.4\textwidth]{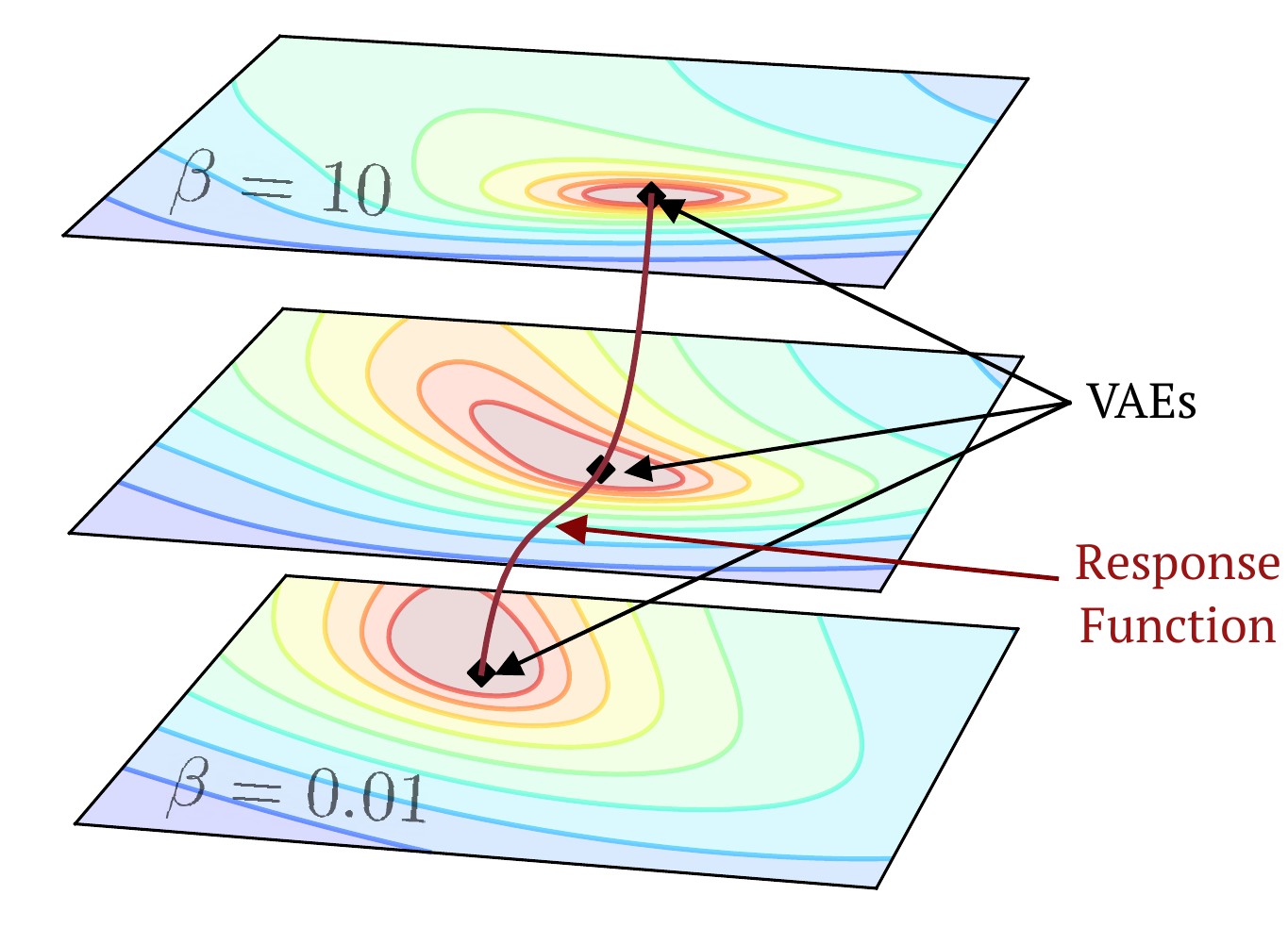}
    \vspace{-0.65cm}
    \small
    \caption{Instead of training several VAEs for each desired KL weight $\beta$, MR-VAEs learn the response functions with a hypernetwork in a single training run.}
    \label{fig:response_function_diagram}
\end{wrapfigure}

We propose explicitly modeling response functions with hypernetworks to construct the rate-distortion curve, as opposed to training multiple VAEs with different KL weights $\beta$. The formulation of response hypernetworks, shown in Figure~\ref{fig:response_function_diagram}, is advantageous as we can infer the optimal parameters for some $\beta$ by simply changing the inputs to the hypernetwork. As response functions map a scalar $\mathbb{R}$ to a high-dimensional parameter space $\mathbb{R}^{m+p}$, there are many possible architectural designs for such response hypernetworks.

Our Multi-Rate VAEs (MR-VAEs) construct a compact approximation of response functions by scaling each layer's weights and bias by a function of the KL weight. More formally, consider the $i$-th layer of a VAE, whose weight and bias are expressed as $\mathbf{W}^{(i)} \in \mathbb{R}^{m_{i+1} \times m_{i}}$ and $\mathbf{b}^{(i)} \in \mathbb{R}^{m_{i+1}}$, respectively. We directly model response functions with hypernetworks $\rvpsi \in \mathbb{R}^h$ by parameterizing each weight and bias as follows\footnote{We show the hypernetwork parameterization for convolution layers in Appendix~\ref{app:implementation_details}.}:
\begin{equation}
  \begin{gathered}
    \mathbf{W}_{\rvpsi}^{(i)} (\beta) =  \sigmoid^{(i)} \left( \mathbf{w}_{\text{hyper}}^{(i)} \log(\beta) + \mathbf{b}_{\text{hyper}}^{(i)}\right) \odot_{\text{row}} \mathbf{W}_{\text{base}}^{(i)}\\
    \mathbf{b}^{(i)}_{\rvpsi} (\beta) = \sigmoid^{(i)} \left( \mathbf{w}_{\text{hyper}}^{(i)} \log(\beta) + \mathbf{b}_{\text{hyper}}^{(i)}\right) \odot
    \mathbf{b}^{(i)}_{\text{base}},
  \end{gathered}
  \label{eq:vae_response_hypernetwork}
\end{equation}
where $\odot$ and $\odot_{\text{row}}$ indicate elementwise multiplication and row-wise rescaling. Here, $\mathbf{w}^{(i)}_{\text{hyper}}, \mathbf{b}^{(i)}_{\text{hyper}} \in \mathbb{R}^{m_{i+1}}$ are vectors that are used in applying an affine transformation to the log KL weight. We further define the elementwise activation function $\sigma^{(i)}(\cdot)$ as:
\begin{align}
    \sigma^{(i)} (x) = \begin{dcases*}
    \frac{1}{1 + e^{-x}} &\text{if $i$-th layer belongs to encoder $\rvtheta$.}\\
    (\text{ReLU}(1 - \exp(x)))^{1/2} &\text{if $i$-th layer belongs to decoder $\rvphi$.}
    \end{dcases*}
    \label{eq:vae_response_activation}
\end{align}
This specific choice of activation functions recovers the exact response function for linear VAEs, which we elaborate on in Section~\ref{subsec:linear_vaes_response_function}. We further note that the above activation functions scale the base weight and bias between 0 and 1. 

Observe that the hypernetwork parameterization in Eqn.~\ref{eq:vae_response_hypernetwork} is both memory and cost-efficient: it needs $2m_{i+1}$ additional parameters to represent the weight and bias and requires 2 additional elementwise multiplications per layer in the forward pass. Given activations $\mathbf{a}^{(i-1)} \in \mathbb{R}^{m_i}$ from the previous layer, the resulting pre-activations $\mathbf{s}^{(i)}$ can alternatively be expressed as:
\begin{align}
    \mathbf{s}^{(i)} &= \mathbf{W}^{(i)}_{\rvpsi} (\beta) \mathbf{a}^{(i-1)} + \mathbf{b}^{(i)}_{\rvpsi} (\beta) \\
    &= \sigmoid^{(i)} \left( \mathbf{w}_{\text{hyper}}^{(i)} \log(\beta) + \mathbf{b}_{\text{hyper}}^{(i)}\right) \odot \left(\mathbf{W}^{(i)}_{\text{base}} \mathbf{a}^{(i-1)} + \mathbf{b}^{(i)}_{\text{base}} \right) \nonumber.
\label{eq:vae_response_hypernetwork_activation}
\end{align}
The architecture for MR-VAEs is illustrated in Figure~\ref{fig:mr_vae_diagram}. This response hypernetwork is straightforward to implement in popular deep learning frameworks (e.g., \texttt{PyTorch}~\citep{paszke2019pytorch} and \texttt{Jax}~\citep{jax2018github}) by replacing existing modules with pre-activations gated modules. We provide a sample \texttt{PyTorch} code in Appendix~\ref{code:pytorch}.

\subsection{Exact Response Functions for linear VAEs}
\label{subsec:linear_vaes_response_function}

Here, we justify the proposed hypernetwork parameterization introduced in Section~\ref{subsec:response_hypernetwork} by analyzing the exact response functions for linear VAEs~\citep{lucas2019don}. We show that MR-VAEs can precisely express the response functions for this simplified model. Consider the following problem (analogous to the setup introduced in Section~\ref{subsec:linear_vae}): 
\begin{equation}
  \begin{gathered}
    p_{\rvtheta}(\bx | \bz) = \mathcal{N}(\mathbf{D}^{(2)}\mathbf{D}^{(1)} \bz + \bmu, \sigma^2 \mathbf{I})\\
    q_{\rvphi}(\bz | \bx) = \mathcal{N}(\mathbf{E}^{(2)} \mathbf{E}^{(1)} (\bx - \bmu), \mathbf{C}^{(2)} \mathbf{C}^{(1)}).
  \end{gathered}
    \label{eq:two_layer_linear_vae}
\end{equation}
Compared to the linear VAE setup in Eqn.~\ref{eq:linear_vae_setup}, we represent the encoder weight, diagonal covariance, and decoder weight as the product of two matrices for compatibility with our framework, where $\mathbf{E}^{(1)}, \mathbf{C}^{(1)}$, and $\mathbf{D}^{(1)}$ are all square matrices. This decomposition can be understood as constructing two-layer linear networks for both encoder and decoder. 

We apply the MR-VAE formulation (shown in Eqn.~\ref{eq:vae_response_hypernetwork}) to all parameters and construct a response hypernetwork for the encoder weight, diagonal covariance, and decoder weight. The following theorem shows that the MR-VAE architecture is capable of representing the response functions for linear VAEs.

\theoremstyle{plain}
\begin{theorem}
Consider the linear VAE model introduced in Eqn.~\ref{eq:two_layer_linear_vae}. Let the encoder weight, diagonal covariance matrix, and decoder weight be parameterized as $\mathbf{E}_{\rvpsi}(\beta) = \mathbf{E}^{(2)}_{\rvpsi} (\beta) \mathbf{E}^{(1)}_{\rvpsi} (\beta)$, $\mathbf{C}_{\rvpsi}(\beta) = \mathbf{C}^{(2)}_{\rvpsi} (\beta) \mathbf{C}^{(1)}_{\rvpsi} (\beta)$, and $\mathbf{D}_{\rvpsi}(\beta) = \mathbf{D}^{(2)}_{\rvpsi} (\beta) \mathbf{D}^{(1)}_{\rvpsi} (\beta)$, respectively. Then, there exist hypernetwork parameters $\rvpsi^{\star} \in \mathbb{R}^h$ such that $\mathbf{E}_{\rvpsi^\star}(\beta), \mathbf{C}_{\rvpsi^\star}(\beta)$, and $\mathbf{D}_{\rvpsi^\star}(\beta)$ are the exact response functions for linear VAEs.
\end{theorem}

The proof is shown in Appendix~\ref{app:theorem1_proof}. 
Note that we use a separate activation function (described in Eqn.~\ref{eq:vae_response_activation}) for encoder and decoder parameters to exactly represent the response functions.
Motivated by the analysis presented in this section, MR-VAEs use the same hypernetwork parameterization for deep VAEs.

\subsection{Training Objective for Response Hypernetwork}

To learn the optimal parameters for various ranges of the KL weight $\beta$, we propose to use an objective analogous to the Self-Tuning Networks (STN) objective~\citep{lorraine2018stochastic,mackay2019self,bae2020delta}. The hypernetwork parameters $\rvpsi$ are optimized to minimize the following objective:
\begin{align}
    \rvpsi^{\star} = \argmin_{\rvpsi \in \mathbb{R}^h} \mathcal{Q}(\rvpsi) = \argmin_{\rvpsi \in \mathbb{R}^h} \mathbb{E}_{\eta \sim \mathcal{U}[\log(a), \log(b)]} \left[ \loss_{\exp(\eta)} (\rvphi_{\rvpsi} (\exp(\eta)), \rvtheta_{\rvpsi} (\exp(\eta))) \right],
    \label{eq:mr_vae_objective}
\end{align}
where $\mathcal{U}[\log(a), \log(b)]$ is an uniform distribution with range $\log(a)$ and $\log(b)$. The proposed objective encourages response hypernetworks to learn optimal parameters for all KL weights $\beta$ in the range between $a$ and $b$. 
Unlike the STN objective, we sample the hypernetwork inputs from a uniform distribution with a fixed range (instead of a Gaussian distribution with a learnable covariance matrix) as we are interested in learning the global response function (instead of the local response function). 

\subsection{Training Algorithm}

\setlength{\textfloatsep}{5pt}
\begin{algorithm*}[t]
    \caption{Multi-Rate Variational Autoencoders (MR-VAEs)}
    \begin{algorithmic}[l]
    \State \textbf{Require:} $\rvpsi$ (Hypernetwork parameters), $\eta$ (learning rate), $(a, b)$ (sample range)
    \While{not converged}
        \State $\batch \sim \Dtrain$  \algorithmiccomment{Sample a mini-batch}
        \State ${\color{dkred} \boldsymbol{\eta}} \sim \mathcal{U}(\log(a), \log(b))$ \algorithmiccomment{Sample inputs to the hypernetwork}
        \State $\mathcal{Q}(\rvpsi) := \loss_{\exp({\color{dkred} \boldsymbol{\eta}})} (\rvphi_{\rvpsi} (\exp({\color{dkred} \boldsymbol{\eta}})), \rvtheta_{\rvpsi} (\exp({\color{dkred} \boldsymbol{\eta}})); \mathcal{B})$ \algorithmiccomment{Compute the MR-VAE objective}
        \State $\rvpsi \gets \rvpsi - \eta \nabla_{\rvpsi} \mathcal{Q}(\rvpsi)$ \algorithmiccomment{Update hypernetwork parameters}
    \EndWhile
    \end{algorithmic}
    \label{alg:mr_vae}
\end{algorithm*}

The full algorithm for MR-VAE is summarized in Alg.~\ref{alg:mr_vae}. When training MR-VAEs, we approximate the objective in Eqn.~\ref{eq:mr_vae_objective} by sampling $\eta$ in each gradient descent iteration. 

\paragraph{Normalization.} 
As it is widely considered beneficial for optimization to normalize the inputs and activations~\citep{lecun2012efficient,montavon2012deep,ioffe2015batch,bae2020delta}, we standardize the inputs to the hypernetworks (after $\log$ transform) to have a mean of 0 and standard deviation of 1. Since the inputs are sampled from a fixed distribution $
\mathcal{U}[\log(a), \log(b)]$, we can apply a fixed transformation based on the sample range $[a, b]$ specified before training.

\paragraph{Memory and Computation Costs.} As detailed in Section~\ref{subsec:response_hypernetwork}, MR-VAE architectures introduce 2 additional vectors for each layer of the base neural network. During the forward pass, MR-VAEs require 2 additional elementwise operations in each layer. Across all our image reconstruction experiments, MR-VAEs require at most a 5\% increase in parameters and wall-clock time compared to $\beta$-VAEs.

\paragraph{Hyperparameters.} While MR-VAE does not require hyperparameter tuning on the KL weight, it needs two hyperparameters $a$ and $b$ that define the sample range for the KL weight. However, we show that MR-VAEs are robust to the choices of these hyperparameters in Section~\ref{subsec:experiment_sensitivity} and we use a fixed value  $a = 0.01$ and $b = 10$ for our image and text reconstruction experiments.

\section{Related Works}
\label{sec:related_works}
\vspace{-0.2cm}

\paragraph{Rate-distortion with VAEs.} 
\citet{alemi2018fixing} and \citet{huang2020evaluating} advocate for reporting points along the rate-distortion curve rather than just the ELBO objective to better  characterize the representation quality of the latent code. For instance, a powerful decoder can ignore the latent code and still obtain high marginal likelihood as observed by \citet{bowman2015generating} and \citet{chen2016variational}. \citet{alemi2018fixing} show that this problem can be alleviated by choosing $\beta < 1$, but this approach still requires training a VAE for each desired information theoretic tradeoff. \citet{yang2020variable} study the use of modulating activations of autoencoders for image compression, but unlike MR-VAEs, they do not learn a generative model and also require learning a separate network for each target rate.
\paragraph{Calibrated Decoders.} Instead of using a fixed decoder variance, calibrated decoders~\citep{lucas2019don,rybkin2020sigmavae} update the decoder variance during training and do not require tuning the KL weight $\beta$. In the case where Gaussian decoders are used, the decoder variance $\sigma^2$ is equivalent to setting $\beta = 2 \sigma^2$. Calibrated decoders may be desirable for applications such as density modeling as it is trained with the standard ELBO objective ($\beta = 1$). However, many applications of VAEs involve directly tuning $\beta$. In addition, calibrated decoders cannot construct a rate-distortion curve and may not be appealing when the aim is to get better disentanglement. MR-VAEs can be directly applied in both of these settings.

\paragraph{Multiplicative Interactions.} Multiplicative interactions have a long history in neural networks. Earlier works include LSTM cells~\citep{hochreiter1997long}, which use gating mechanisms to model long-term memories, and Mixture of Experts~\citep{jacobs1991adaptive}, which selects learners with a softmax gating. The FiLM network~\citep{perez2018film} and conditional batch normalization~\citep{perez2017learning,brock2018large} further extend the gating mechanism by scaling and shifting the pre-activations. \citet{jayakumar2020multiplicative} analyze the expressive power of multiplicative interactions within a network. Our work uses multiplicative interactions to learn the response function for $\beta$-VAEs. 

\paragraph{Response Hypernetworks.} Our use of hypernetworks for learning best response functions is related to Self-Tuning Networks (STNs)~\citep{mackay2019self,bae2020delta} and Pareto Front Hypernetworks (PFHs)~\citep{navon2020learning}. \citet{lorraine2018stochastic} first constructed response hypernetworks for bilevel optimization~\citep{colson2007overview} and optimized both parameters and hyperparameters in a single training run. To support high dimensional hypernetwork inputs, STNs use structured response hypernetworks similar to MR-VAEs, whereas PFHs use chunking that iteratively generates network parameters. In contrast to these works, the architecture of MR-VAEs is specifically designed for learning $\beta$-VAE's response functions and requires significantly less computation and memory. On a similar note, \citet{dosovitskiy2019you} also proposed an efficient method to approximate the response functions using the FiLM parameterization and demonstrated the effectiveness of modeling $\beta$-VAE's response function. 

\vspace{-0.2cm}
\section{Experiments}
\label{sec:experiments}
\vspace{-0.3cm}

Our experiments investigate the following questions:
(1) Can MR-VAEs learn the optimal response functions for linear VAEs?
(2) Does our method scale to modern-size $\beta$-VAEs? (3) How sensitive are MR-VAEs to hyperparameters?
(4) Can MR-VAEs be applied to other VAE models such as $\beta$-TCVAEs~\citep{chen2018isolating}? 

We trained MR-VAEs to learn VAEs with multiple rates in a single training run on image and text reconstruction tasks using several network architectures, including convolutional-based VAEs, autoregressive VAEs, and hierarchical VAEs. 
In addition, we applied MR-VAEs to the $\beta$-TCVAE objective to investigate if our method can be extended to other models and help find VAEs with improved disentangled representations. Experiment details and additional experiments, including ablation studies, are provided in Appendix~\ref{app:experiment_details} and Appendix~\ref{app:additional_experiments}, respectively.  

\subsection{How do MR-VAEs perform on Linear VAEs?}

\begin{wrapfigure}[13]{R}{0.5\textwidth}
    \centering
    \vspace{-1.cm}
    \includegraphics[width=0.44 \textwidth]{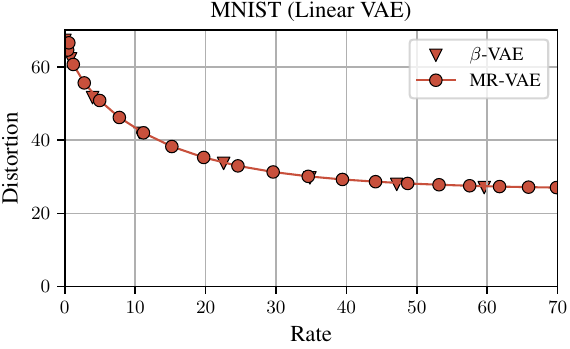}
    \vspace{-0.2cm}
    \small
    \caption{A comparison of MR-VAEs and $\beta$-VAEs on linear VAEs. MR-VAEs learn the optimal rate-distortion curve in a single training run.}
    \label{fig:linear_vae_rd_curve}
\end{wrapfigure}

We first considered two-layer linear VAE models (introduced in Section~\ref{subsec:linear_vaes_response_function}) to validate our theoretical findings. We trained MR-VAEs on the MNIST dataset~\citep{deng2012mnist} by sampling the KL weighting term $\beta$ from 0.01 to 10. Then, we trained 10 separate two-layer linear $\beta$-VAEs each with different $\beta$ values. 

The rate-distortion curve for MR-VAEs and $\beta$-VAEs are shown in Figure~\ref{fig:linear_vae_rd_curve}. We observed that the rate-distortion curve generated with MR-VAEs closely aligns with the rate-distortion curve generated with individual retraining. Furthermore, by explicitly formulating response functions, MR-VAEs do not require retraining a VAE for each $\beta$. 

\subsection{Can MR-VAEs scale to modern-size Architectures?}
\label{subsec:modern_experiments}

\begin{figure}[!t]
    \vspace{-0.4cm}
    \centering
    \begin{tabular}{ccc}
    \includegraphics[width=0.32\linewidth]{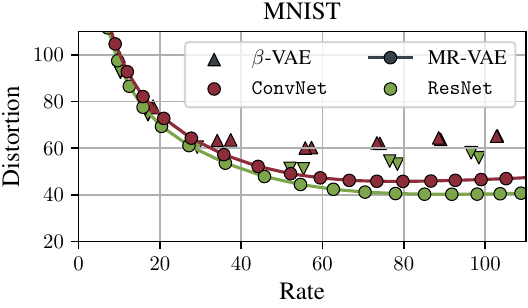}
    \includegraphics[width=0.32\linewidth]{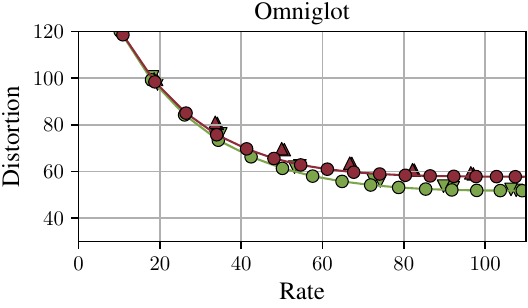}
    \includegraphics[width=0.32\linewidth]{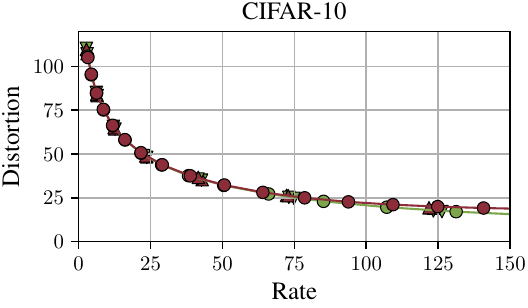}\\
    \includegraphics[width=0.32\linewidth]{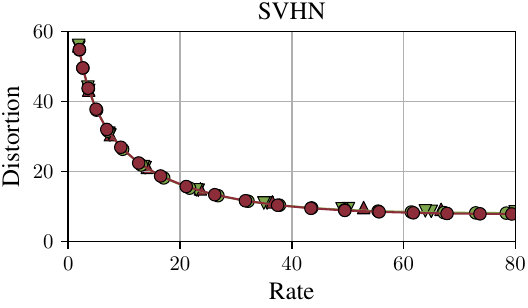}
    \includegraphics[width=0.32\linewidth]{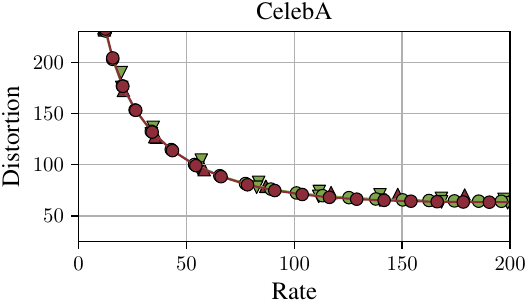}
    \includegraphics[width=0.32\linewidth]{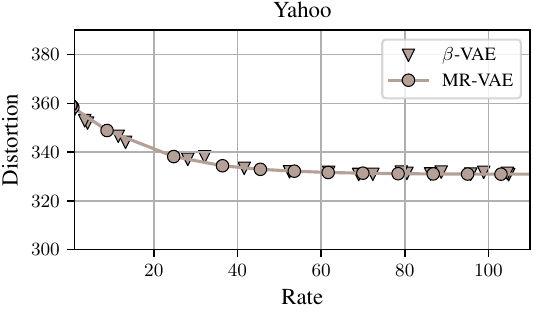}\\
    \includegraphics[width=0.32\linewidth]{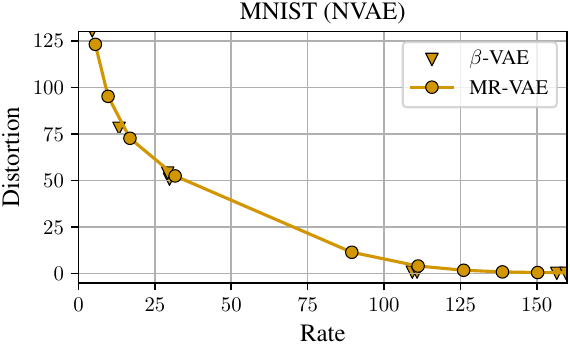}
    \includegraphics[width=0.32\linewidth]{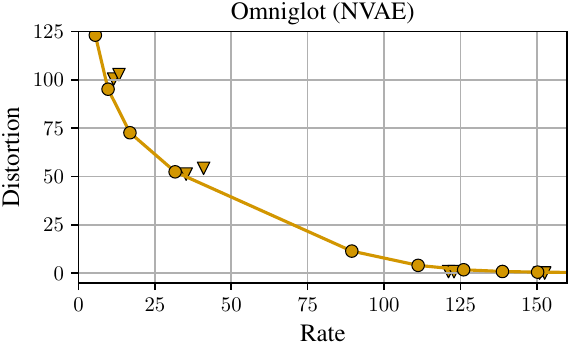}
    \includegraphics[width=0.32\linewidth]{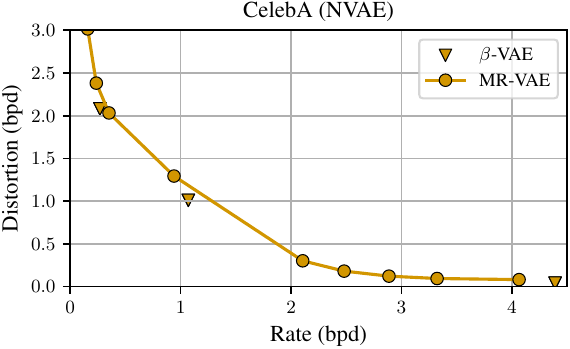}
    \end{tabular}
    \vspace{-0.1cm}
    \caption{\small A comparison of MR-VAEs and $\beta$-VAEs (with and without KL annealing). $\beta$-VAEs were trained multiple times on several KL weighting $\beta$ to construct the rate-distortion curve. Across all tasks, MR-VAEs achieve a competitive rate-distortion curve with $\beta$-VAEs on a single run of training.}
    \label{fig:image_rd_curve}
    \vspace{0.1cm}
\end{figure}

We trained convolution and ResNet-based architectures on binary static MNIST~\citep{larochelle2011neural}, Omniglot~\citep{lake2015human}, CIFAR-10~\citep{krizhevsky2009learning}, SVHN~\citep{netzer2011reading}, and CelebA64~\citep{liu2015faceattributes} datasets, following the experimental set-up from~\citet{chadebec2022pythae}. We also trained NVAEs on binary static MNIST and Omniglot datasets using the same experimental set-up from~\citet{vahdat2020NVAE}. Lastly, we trained autoregressive LSTM VAEs on the Yahoo dataset with the set-up from~\citet{he2019lagging}.

We sampled the KL weighting term from 0.01 to 10 for both MR-VAEs and $\beta$-VAEs. Note that the training for $\beta$-VAEs was repeated 10 times (4 times for NVAEs) with log uniformly spaced $\beta$ to estimate a rate-distortion curve. We compared MR-VAEs with individually trained $\beta$-VAEs and show test rate-distortion curves in Figure~\ref{fig:image_rd_curve}. Note that traditional VAEs typically require scheduling the KL weight to avoid posterior collapse ~\citep{bowman2015generating, fu2019cyclical, lucas2019don}, so we train $\beta$-VAEs with both a constant and a KL annealing schedule.
Across all tasks, MR-VAEs achieved competitive rate-distortion curves with the independently trained VAEs and even improved performance on the MNIST, Omniglot, and CelebA datasets. 
Note that we focus on the visualization of the rate-distortion curve as advocated for by \citet{alemi2016deep, huang2020evaluating}.

\begin{figure}[!t]
    \centering
    \begin{tabular}{ccc}
    \includegraphics[width=0.32\linewidth]{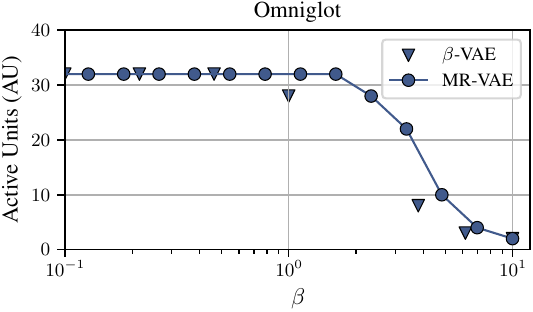}
    \includegraphics[width=0.32\linewidth]{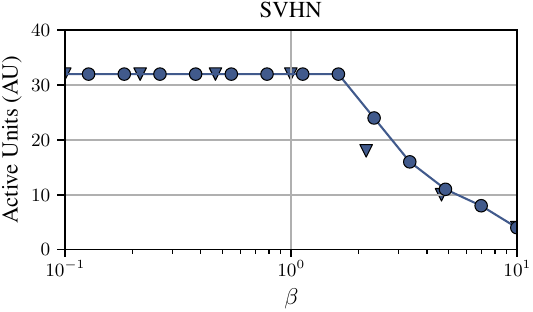}
    \includegraphics[width=0.32\linewidth]{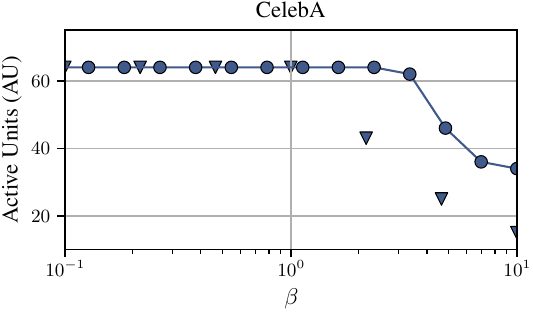}\\
    \includegraphics[width=0.32\linewidth]{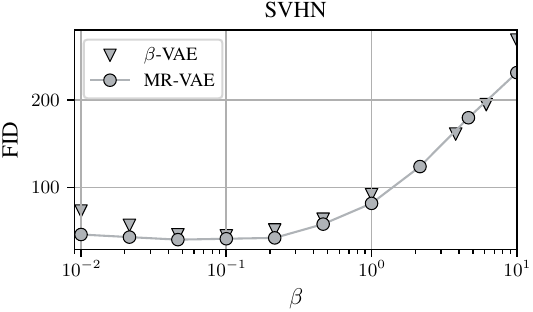}
    \includegraphics[width=0.32\linewidth]{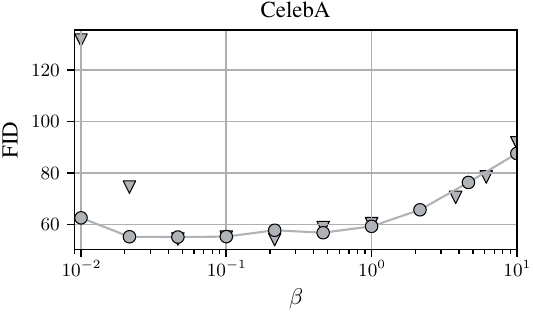}
    \end{tabular}
    \vspace{-0.1cm}
    \caption{\small A comparison of independently trained $\beta$-VAEs (with KL annealing) and MR-VAEs on proxy metrics such as \textbf{(top)} active units (AU) and \textbf{(bottom)} Fréchet inception distance (FID). Note that a lower FID is better.}
    \label{fig:image_other_curve}
\end{figure}

\begin{figure*}[t]
    \small
    \centering
    \vspace{-0.4cm}
    \includegraphics[width=\linewidth]{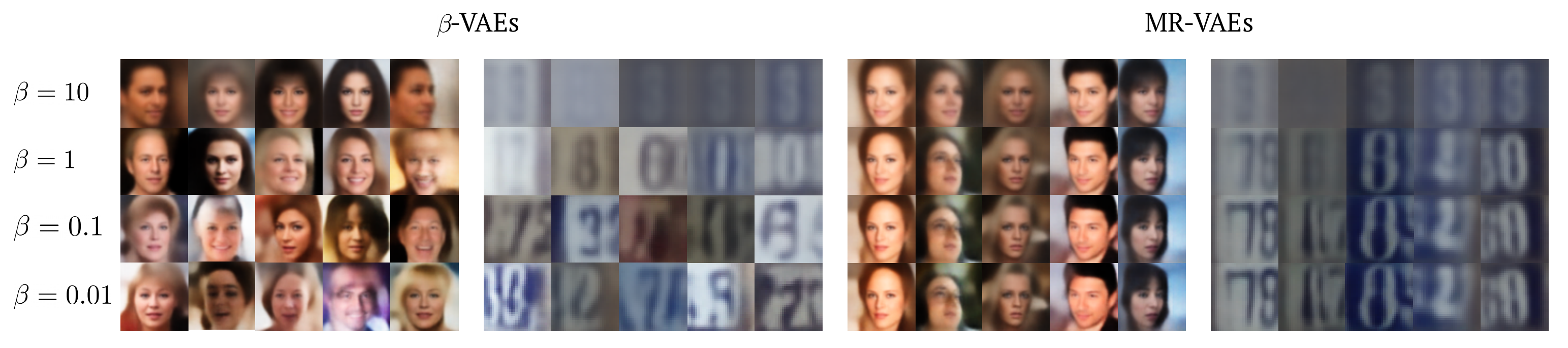}
    \caption{Generative samples for $\beta$-VAEs and MR-VAEs on CelebA and SVHN datasets. We use the same latent variables $\mathbf{z}$ sampled from the prior for all columns. For MR-VAEs, there is consistency among the generated images in each column since intermediate weights are shared for different $\beta$. While higher $\beta$ values result in a worse reconstruction loss for both models, MR-VAEs produce images that are of higher quality.
    }
    \label{fig:mr_vae_samples}
\end{figure*}

As MR-VAEs directly learn the optimal parameters corresponding to some $\beta$ in a single training run, we can use them analyze the dependency between the KL weight $\beta$ and various proxy metrics with a single training run. To demonstrate this capability of MR-VAEs, we compute the Fréchet Inception Distance (FID)~\citep{heusel2017gans} for natural images and active units (AU)~\citep{burda2016iwae} in Figure~\ref{fig:image_other_curve}. Notably, by using a shared structure for various KL weights, we observed that MR-VAEs are more resistant to dead latent variables with high KL weights are used and generate more realistic images with low KL weights. 

We further show samples from both $\beta$-VAEs and MR-VAEs on SVHN and CelebA datasets which used ResNet-based encoders and decoders in Figure~\ref{fig:mr_vae_samples}. One interesting property is that MR-VAEs have consistency among the generated images for the same sampled latent since base VAE weights are shared for different $\beta$.
A practitioner can select the desired model for their task or trade-off between compression and generation quality by feeding different $\beta$s into the hypernetwork at inference time.

\subsection{How sensitive are MR-VAEs to hyperparameters?}
\label{subsec:experiment_sensitivity}
\begin{figure}[!t]
    \centering
    \begin{tabular}{ccc}
    \includegraphics[width=0.32\linewidth]{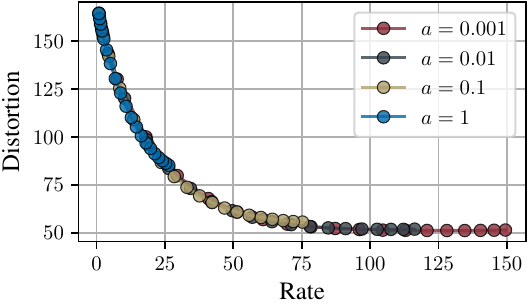}
    \includegraphics[width=0.32\linewidth]{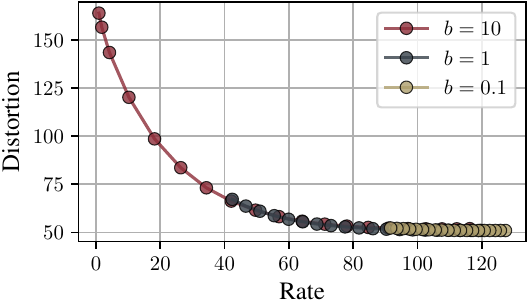}
    \includegraphics[width=0.32\linewidth]{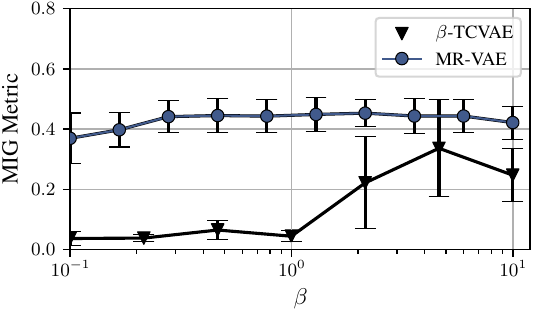}
    \end{tabular}
    \vspace{-0.1cm}
    \caption{\small \textbf{(left and middle)} An ablation studying the effect of the sample range for MR-VAEs. For each figure, we fix one value ($b = 10$ or $a = 0.01$) and change the other. \textbf{(right)} A comparison of independently trained $\beta$-TCVAEs and MR-VAEs on the Mutual Information Gap (MIG) metric (the higher value is better)}
    \label{fig:sensitivity}
\end{figure}

While the MR-VAE eliminates the need to train the network multiple times with different KL weights and a KL schedule, it introduces two hyperparameters $a$ and $b$ that determine the range to sample the KL weight. Here, we show that MR-VAEs are insensitive and robust to the choices of the hyperparameters and can be fixed through various applications. We trained ResNet encoders and decoders on the Omniglot dataset with the same configuration as Section~\ref{subsec:modern_experiments} and show the test rate-distortion curves in Figure~\ref{fig:sensitivity}. In the left, we fixed $b = 10$ and changed the sample range $a$ in the set \{0.001, 0.01, 0.1, 1\} and in the middle, we fixed $a = 0.01$ and modified the sample range $b$ in the set \{10, 1, 0.1\}. While the length of the rate-distortion curves differs with different sample ranges, we observed that the area under the curve is similar across all configurations.

\subsection{Can MR-VAEs be Extended to Other Models?}

To demonstrate the applicability of MR-VAEs to other VAE models, we trained MR-VAEs on the $\beta$-TCVAE objective. The weight in $\beta$-TCVAE balances the reconstruction error and the total correlation, instead of the reconstruction error and the KL divergence term.

We trained MR-VAEs composed of MLP encoders and decoders on the dSprites dataset, following the set-up from~\citet{chen2018isolating}. 
Since we are interested in the disentanglement ability of the model, we sampled the weighting $\beta$ between 0.1 and 10. We compare $\beta$-TCVAEs and MR-VAEs (trained with the $\beta$-TCVAE objective) by examining their performance on the Mutual Information Gap (MIG) in Figure~\ref{fig:sensitivity}. We observed that MR-VAEs are more robust to the choice of the weighting term and achieve competitive final results as the baseline $\beta$-TCVAEs without having to train multiple models.

\section{Conclusion}
In this work, we introduce Multi-Rate VAE (MR-VAE), an approach for learning multiple VAEs with different KL weights in a single training run. Our method eliminates the need for extensive hyperparameter tuning on the KL weight, while only introducing a small memory and computational overhead. 
The key idea is to directly learn a map from KL weights to optimal encoder and decoder parameters using a hypernetwork. On a variety of tasks and architectures, we show that MR-VAEs generate competitive or even better rate-distortion curves compared to the baseline of retraining multiple VAEs with various values of $\beta$. Moreover, our framework is general and is straightforward to extend to various existing VAE architectures.

\section*{Acknowledgements}
We would like to thank James Lucas, Rob Brekelmans, Xuchan Bao, Nikita Dhawan, Cem Anil, Silviu Pitis, and Alexey Dosovitskiy for their valuable feedback on this paper. We would also like to thank Sicong Huang and many other colleagues for their helpful discussions throughout this research. Resources used in this research were provided, in part, by the Province of Ontario, the Government of Canada through CIFAR, and companies sponsoring the Vector Institute (\url{www.vectorinstitute.ai/partners}).

\bibliography{main}
\bibliographystyle{icml2022}

\newpage
\appendix

\section{Derivations}
\label{app:derivation}

\subsection{Analytic Solutions to Linear VAEs}
\label{subapp:linear_vae_analysis}

In this section, we present the analysis from~\citet{lucas2019don} with some additional details. Recall that the linear VAE is defined as:
\begin{equation}
  \begin{gathered}
    p_{\rvtheta}(\bx | \bz) = \mathcal{N}(\mathbf{D} \bz + \bmu, \sigma^2 \mathbf{I})\\
    q_{\rvphi}(\bz | \bx) = \mathcal{N}(\mathbf{E} (\bx - \bmu), \mathbf{C}),
  \end{gathered}
   \label{eq:linear_vae_setup2}
\end{equation}
where $\mathbf{D}$ is the decoder weight, $\mathbf{E}$ is the encoder weight, and $\mathbf{C}$ is the diagonal covariance matrix. Also recall that the $\beta$-VAE objective is defined as:
\begin{align}
    \loss_{\beta} (\rvphi, \rvtheta) =  \mathbb{E}_{p_d (\mathbf{x})}[\mathbb{E}_{q_{\rvphi}(\bz | \bx)}[-\log p_{\rvtheta} (\bx| \bz)]] + \beta \mathbb{E}_{p_d (\mathbf{x})} [\dkl(q_{\rvphi}(\bz | \bx), p(\bz))].
    \label{eq:beta_vae_objective2}
\end{align}
For linear VAEs, each component of the $\beta$-VAE objective can be expressed in closed form. For simplicity, we assume a fixed identity covariance matrix $\mathbf{I}$ for the decoder as modifying $\beta$ gives a similar effect as changing the observation noise~\citep{rybkin2020sigmavae}. We also assume that the data covariance matrix is full rank. The second KL term can be expressed as:
\begin{align}
    \dkl(q_{\rvphi}(\bz|\bx), p(\bz)) &= \frac{1}{2} \left( -\log\det \mathbf{C} + (\bx - \bmu)^\top \mathbf{E}^\top \mathbf{E} (\bx - \bmu) + \trace(\mathbf{C}) - d \right).
\end{align}
The first term can be expressed as:
\begin{align}
    \mathbb{E}_{q_{\rvphi}(\bz|\bx)} &[\log p_{\rvtheta}(\bx|\bz)] = \mathbb{E}_{q_{\rvphi}(\bz|\bx)} \left[ - (\mathbf{D} \bz - (\bx - \bmu))^\top (\mathbf{D} \bz - (\bx - \bmu)) / 2 - c \right] \\
    &= \mathbb{E}_{q_{\rvphi}(\bz|\bx)} \left[ \frac{- (\mathbf{D}\bz)^\top (\mathbf{D}\bz) + 2 (\bx - \bmu)^\top \mathbf{D} \bz - (\bx - \bmu)^\top (\bx - \bmu)}{2} - c \right], 
\end{align}
where $c = \sfrac{d}{2} \log 2 \pi$ is a constant term. Because $\mathbf{D}\bz \sim \mathcal{N}(\mathbf{D}\mathbf{E} (\bx - \bmu), \mathbf{D}\mathbf{C}\mathbf{D}^\top)$, we can further simplify the equation as:
\begin{align}
    \mathbb{E}_{q_{\rvphi}(\bz|\bx)} [\log p_{\rvtheta}(\bx|\bz)] &= \frac{1}{2} (-\trace(\mathbf{D}\mathbf{C}\mathbf{D}^\top) - (\bx - \bmu)^\top \mathbf{E}^\top \mathbf{D}^\top \mathbf{D} \mathbf{E} (\bx - \bmu) \\
    &\quad + 2(\bx - \bmu)^\top \mathbf{W} \mathbf{V} (\bx - \bmu) - (\bx - \bmu)^\top (\bx - \bmu)) - c.
\end{align}

After setting $\bmu = \bmu_{\text{MLE}}$ (which is the global minimum for $\bmu$ as discussed in \citet{lucas2019don}), we take the gradient with respect to $\mathbf{C}$ and $\mathbf{E}$ to compute critical points of encoder weights and covariance matrix. These gradients can be written as:
\begin{align}
    \frac{\partial \loss_{\beta}}{\partial \mathbf{C}} (\vphi, \rvtheta) &= \frac{N}{2} \left( \beta \mathbf{C}^{-1} - \beta \mathbf{I} - \text{diag}(\mathbf{D}^\top \mathbf{D})\right). \\
    \frac{\partial \loss_{\beta}}{\partial \mathbf{E}} (\vphi, \rvtheta)&= N \left( \mathbf{D}^\top \mathbf{S} - \mathbf{D}^\top \mathbf{D} \mathbf{E} \mathbf{S} - \beta \mathbf{E} \mathbf{S} \right),
\end{align}
where $\mathbf{S}$ is the sample covariance matrix. By setting these gradients to equal to $\mathbf{0}$, the critical points can be expressed as:
\begin{align}
    \mathbf{C}^\star &= \beta ( \text{diag}(\mathbf{D}^\top \mathbf{D}) + \beta \mathbf{I} )^{-1}.\\
    \mathbf{E}^\star &= (\mathbf{D}^\top \mathbf{D} + \beta \mathbf{I})^{-1} \mathbf{D}^\top,
\end{align}
which recovers the true posterior mean and posterior covariance when the columns of $\mathbf{D}$ are orthogonal. Moreover, as the global minimum of the decoder weight coincides with the solution to the probabilistic PCA (pPCA)~\citep{dai2018connections,lucas2019don,sicks2021generalised}, we can express the optimal decoder weight as:
\begin{align}
    \mathbf{D}^\star = \mathbf{U} (g(\boldsymbol{\Lambda} - \beta \mathbf{I}))^{\sfrac{1}{2}} \mathbf{R},
\end{align}
where $\mathbf{U}$ corresponds to the first $k$ eigenvectors of the sample covariance $\mathbf{S}$ with corresponding top eigenvalues $\lambda_1 \geq ... \geq \lambda_k$ stored in the $k \times k$ diagonal matrix $\boldsymbol{\Lambda}$. The matrix $\mathbf{R}$ is some rotation matrix and the function $g(x) = \max(0, x)$ clips negative values to 0 and is applied elementwise.

\subsection{Proof of Theorem 1}
\label{app:theorem1_proof}

In this section, we show that the MR-VAE parameterization can represent the response functions for linear VAEs. Throughout this section, we use $\eta = \log \beta$ for simplicity in notation. First, notice that we can simplify the equation for the encoder response function by using the singular value decomposition $\mathbf{D} = \mathbf{A} \rmS \rmB^\top$:
\begin{align}
    \mathbf{E}^\star (\eta)
    &= \mathbf{B} (\mathbf{S}^2  + \exp(\eta) \mathbf{I})^{-1} \mathbf{S} \mathbf{A}^\top.
\end{align}
Suppose we apply the MR-VAE parameterization to two-layer encoders $\mathbf{E}_{\rvpsi}(\eta) = \mathbf{E}^{(2)}_{\rvpsi} (\eta) \mathbf{E}^{(1)}_{\rvpsi} (\eta)$, where each component can be expressed as:
\begin{align}
    \mathbf{E}_{\rvpsi}^{(1)}(\eta) &= \sigma_E (\mathbf{v}^{(1)} \eta + \mathbf{s}^{(1)} )\odot_{\text{row}} \mathbf{E}^{(1)}_{\text{base}}\\
    \mathbf{E}_{\rvpsi}^{(2)}(\eta) &= \sigma_E (\mathbf{v}^{(2)} \eta + \mathbf{s}^{(2)} )\odot_{\text{row}} \mathbf{E}^{(2)}_{\text{base}},
\end{align}
where $\sigma_E (\cdot)$ is the sigmoid activation function. Consider setting $\mathbf{E}^{(1)}_{\text{base}} = \mathbf{A}^\top$, $\mathbf{v}^{(1)} = -\mathbf{1}$, and $\mathbf{s}^{(1)}_i = 2\log(\mathbf{S}_{ii})$ for all $i$ in dimension of $\mathbf{s}^{(1)}$. Then, each dimension of the scaling term can be expressed as:
\begin{align}
    \sigma_E (\mathbf{v}^{(1)} \eta + \mathbf{s}^{(1)})_i &= \frac{1}{1 + \exp(\eta - 2\log(\mathbf{S}_{ii}) )} = \frac{\mathbf{S}_{ii}^2}{\mathbf{S}_{ii}^2 + \exp(\eta)}.
\end{align}
By further setting $\mathbf{E}^{(2)}_{\rvpsi} (\eta) = \mathbf{B} \mathbf{S}^{-1}$, which can be achieved by setting both scaling parameters  ($\mathbf{v}^{(2)}$ and $\mathbf{s}^{(2)}$) to $\mathbf{0}$ and setting $\mathbf{E}^{(2)}_{\text{base}} = 2\mathbf{B} \mathbf{S}^{-1}$, we recover the response function for the encoder weight:
\begin{align}
    \mathbf{E}^{(2)}_{\rvpsi} (\eta) \mathbf{E}^{(1)}_{\rvpsi} (\eta) = \mathbf{B} (\mathbf{S}^2  + \exp(\eta) \mathbf{I})^{-1} \mathbf{S} \mathbf{A}^\top.
\end{align}
The response function for the diagonal covariance can be expressed as:
\begin{align}
    \mathbf{C}^\star  (\eta)
    &= \exp(\eta) (\text{diag}(\mathbf{D}^\top \mathbf{D}) + \exp(\eta) \mathbf{I})^{-1}.
\end{align}
Our hypernetwork parameterization corresponds to $\mathbf{C}_{\rvpsi}(\eta) = \mathbf{C}^{(2)}_{\rvpsi} (\eta) \mathbf{C}^{(1)}_{\rvpsi} (\eta)$, where each component can be written as:
\begin{align}
    \mathbf{C}_{\rvpsi}^{(1)}(\eta) &= \sigma_E (\mathbf{t}^{(1)} \eta + \mathbf{p}^{(1)} )\odot_{\text{row}} \mathbf{C}^{(1)}_{\text{base}}\\
    \mathbf{C}_{\rvpsi}^{(2)}(\eta) &= \sigma_E (\mathbf{t}^{(2)} \eta + \mathbf{p}^{(2)} )\odot_{\text{row}} \mathbf{C}^{(2)}_{\text{base}},
\end{align}
We describe the procedure for constructing the diagonal matrix $\mathbf{C}_{\text{base}}^{(1)}$: in the $i$-th dimension, if $\text{diag}(\mathbf{D}^\top \mathbf{D})_{ii} = 0$, we set the corresponding diagonal entry to 2 and  the scaling (hyper) parameters to be $\mathbf{t}_i^{(1)} = \mathbf{p}^{(1)}_i = 0$.
\newline
Otherwise, if $\text{diag}(\mathbf{D}^\top \mathbf{D})_{ii} \neq 0$, we set the diagonal entry to 1.  We set $\mathbf{t}^{(1)}_i = 1$ and $\mathbf{p}^{(1)}_i = -\log(\text{diag}(\mathbf{D}^\top \mathbf{D})_{ii})$ for each dimension. Finally, by letting $\mathbf{C}_{\rvpsi}^{(2)}(\eta) = \mathbf{I}$, we can achieve:
\begin{align}
    \mathbf{C}^{(2)}_{\rvpsi} (\eta) \mathbf{C}^{(1)}_{\rvpsi} (\eta) = \exp(\eta) (\text{diag}(\mathbf{D}^\top \mathbf{D}) + \exp(\eta) \mathbf{I})^{-1}.
\end{align}
Lastly, recall that the response function for decoder weight can be expressed as:
\begin{align}
\mathbf{D}^\star (\eta) &= \mathbf{U} (g(\boldsymbol{\Lambda} - \exp(\eta) \mathbf{I}))^{\sfrac{1}{2}} \mathbf{R}.
\end{align}
For notational simplicity, we represent $\boldsymbol{\Lambda}_{ii} = \exp(\bxi_i)$ for $i = 1, ..., k$ using the assumption above that the data covariance matrix is full rank. Then, the diagonal term in the response function can be represented as:
\begin{align}
    (g(\boldsymbol{\Lambda} - \exp(\eta) \mathbf{I}))^{\sfrac{1}{2}}_{ii} &= \max(0, (\exp(\bxi_i) - \exp(\eta))^{\sfrac{1}{2}}) \\
   &=  \exp(\bxi_i)^{\sfrac{1}{2}} \max(0, 1 - \exp(-\bxi_i)\exp(\eta))^{\sfrac{1}{2}} \\
   &=  \exp(\bxi_i)^{\sfrac{1}{2}} \max(0, 1 - \exp(\eta - \bxi_i))^{\sfrac{1}{2}}.
\end{align}
Again, recall that we can use MR-VAE parameterization $\mathbf{D}_{\rvpsi}(\eta) = \mathbf{D}^{(2)}_{\rvpsi} (\eta) \mathbf{D}^{(1)}_{\rvpsi} (\eta)$. Specifically, we have:
\begin{align}
    \mathbf{D}_{\rvpsi}^{(1)}(\eta) &= \sigma_D (\mathbf{h}^{(1)} \eta + \mathbf{q}^{(1)} )\odot_{\text{row}} \mathbf{D}^{(1)}_{\text{base}}\\
    \mathbf{D}_{\rvpsi}^{(2)}(\eta) &= \sigma_D (\mathbf{h}^{(2)} \eta + \mathbf{q}^{(2)} )\odot_{\text{row}} \mathbf{D}^{(2)}_{\text{base}},
\end{align}
We highlight that we use a different activation function $\sigma_D (\cdot)$ for representing the decoder weight as described in Eqn.~\ref{eq:vae_response_activation}. We propose to set $\mathbf{D}^{(1)}_{\text{base}} = \mathbf{R}$, $\mathbf{h}^{(1)} = \mathbf{1}$, $\mathbf{q}^{(1)}_{i} = -\bxi_i$, and $\mathbf{D}^{(2)}_{\rvpsi} (\eta) = \mathbf{U} \boldsymbol{\Lambda}^{\sfrac{1}{2}}$ for all $i$. Then, the response hypernetwork can be expressed as:
\begin{align}
    \mathbf{D}_{\rvpsi}^{(2)}(\eta) \mathbf{D}_{\rvpsi}^{(1)}(\eta) = \mathbf{U} (g(\boldsymbol{\Lambda} - \exp(\eta) \mathbf{I}))^{\sfrac{1}{2}} \mathbf{R}.
\end{align}
which matches the response function for the decoder weight. 

\section{Additional Implementation Details}
\label{app:implementation_details}

\subsection{Convolution Layers}
In this section, we present the hypernetwork parameterization for convolutional layers. Consider the $i$-th layer of a convolutional VAE with $C_i$ filters and kernel size $K_i$. We denote $\mathbf{W}^{(i,c)} \in \mathbb{R}^{C_{i-1} \times K_i \times K_i}$ and $\mathbf{b}^{(i,c)} \in \mathbb{R}$ to be the weight and bias of the $c$-th filter, where $c \in \{1, ..., C_i\}$. We formulate the response hypernetwork for the weight and bias as:
\begin{align}
    \mathbf{W}_{\rvpsi}^{(i,c)} (\beta) &= \sigma^{(i)} \left(\mathbf{w}_{\text{hyper}}^{(i, c)}  \log \beta + \mathbf{b}_{\text{hyper}}^{(i, c)} \right) \odot \mathbf{W}_{\text{base}}^{(i,c)} \\
    \mathbf{b}_{\rvpsi}^{(i,c)} (\beta) &= \sigma^{(i)} \left(\mathbf{w}_{\text{hyper}}^{(i, c)} \log \beta + \mathbf{b}_{\text{hyper}}^{(i, c)} \right) \odot \mathbf{b}_{\text{base}}^{(i,c)},
\end{align}
where $\mathbf{w}_{\text{hyper}}^{(i, c)}, \mathbf{b}_{\text{hyper}}^{(i, c)} \in \mathbb{R}$. Observe that the proposed hypernetwork parameterization is similar to that of fully-connected layers and only requires $2C_i$ additional parameters to represent the weight and bias. In the forward pass, the convolutional response hypernetwork requires 2 additional elementwise operations. Hence, MR-VAEs parameterization for convolutional layers also incurs a small computation and memory overhead. Note that Self-Tuning Networks~\citep{mackay2019self,bae2020delta} use an analogous hypernetwork parameterization for convolution layers.  

\newpage 
\subsection{PyTorch Implementation}
\label{code:pytorch}
We show the \texttt{PyTorch} implementation of the MR-VAE layer in Listing~\ref{lst:hc_layer}. The MR-VAE layer can be viewed as pre-activations gating and can be introduced after linear or convolutional layers. 

\begin{lstlisting}[language=Python, label={lst:hc_layer}, caption=MR-VAE Layer implemented in \texttt{PyTorch}., captionpos=b]
from torch import nn
import torch

class MRVAELayer(nn.Module):

  def __init__(self, features: int, activation_fnc: nn.Module) -> None:
    super().__init__()

    self.features = features
    self.hyper_block_scale = nn.Linear(1, self.features, bias=True)
    self.activation_fnc = activation_fnc

  def forward(self, inputs: torch.Tensor, betas: torch.Tensor) -> torch.Tensor:
    scale = self.hyper_block_scale(betas)
    scale = torch.activation_fnc(scale)

    if len(inputs.shape) == 4:
      # Unsqueeze for convolutional layers.
      scale = scale.unsqueeze(-1).unsqueeze(-1)

    return scale * inputs
\end{lstlisting}

\subsection{Memory Overhead for MR-VAE}

In Table~\ref{tab:params_requirement}, we show the number of additional parameters MR-VAEs used in our experiment. The memory required for training MR-VAEs is similar to that of training a single $\beta$-VAE model. 

\begin{table}[]
    \centering
    \resizebox{0.85\columnwidth}{!}{%
    \begin{tabular}{@{}ccccc@{}}
    \toprule
    \multirow{2}{*}{\textbf{Dataset}} & \multirow{2}{*}{\textbf{Architecture}} & \multicolumn{2}{c}{\textbf{Number of Parameters}} & \multirow{2}{*}{\textbf{Increase Percentage}} \\ \cmidrule(lr){3-4}
     &  & \textbf{$\beta$-VAE} & \textbf{MR-VAE} &  \\ \midrule
    \multirow{3}{*}{MNIST \& Omniglot} & CNN & 17.53 $\times 10^5$ & 17.88 $\times 10^5$ & 2.04\% \\
     & ResNet & 18.81 $\times 10^4$ & 18.84 $\times 10^4$ & 0.12\% \\
     & NVAE & 88.68 $\times 10^5$ & 89.50 $\times 10^5$ & 0.92\% \\ \midrule
    \multirow{2}{*}{CIFAR \& SVHN} & CNN & 40.46 $\times 10^5$ & 40.81 $\times 10^5$ & 0.88\% \\
     & ResNet & 78.96 $\times 10^4$ & 78.98 $\times 10^4$ & 0.02\% \\ \midrule
    \multirow{2}{*}{CelebA64} & CNN & 34.59 $\times 10^5$ & 34.96 $\times 10^5$ &  1.08\%\\
     & ResNet & 27.70 $\times 10^4$ & 27.74 $\times 10^4$ & 0.13\% \\ \midrule
    Yahoo & LSTM & 88.21 $\times 10^4$ & 93.51 $\times 10^4$  & 6.01\% \\ \bottomrule
    \end{tabular}%
    }
    \caption{An overview of the parameters required by $\beta$-VAE and MR-VAE.}
    \vspace{0.2cm}
    \label{tab:params_requirement}
\end{table}
\section{Experiment Details}
\label{app:experiment_details}

This section describes the details of the experiments presented in Section~\ref{sec:experiments}. All experiments were implemented using PyTorch~\citep{paszke2019pytorch} and Jax~\citep{jax2018github} libraries and conducted with NVIDIA P100 GPUs, except for NVAE models, which were conducted with NVIDIA RTX A5000 GPUs. 

\subsection{Linear VAEs}

We used the MNIST~\citep{deng2012mnist} dataset for training linear VAEs. We used the Adam optimizer~\citep{kingma2014adam} and trained the network for 200 epochs with 10 epochs of learning rate warmup and a cosine learning rate decay. The latent dimension was set to 32. We conducted hyperparameter searches over learning rates \{0.01, 0.003, 0.001, 0.0003, 0.0001, 0.00003, 0.00001\} with $\beta$s uniformly sampled between 0.01 and 10 on a log scale. We selected the learning rate that achieved the lowest average training loss. The experiments were repeated 3 times with different random seeds and the mean is presented.

MR-VAEs were trained using the same training and architectural configurations as the baseline. We performed a grid search on learning rates \{0.01, 0.003, 0.001, 0.0003, 0.0001, 0.00003, 0.00001\} and selected the learning rate that achieved the best training loss at $\beta = 1$. We also repeated the experiments 3 times with different random seeds.

\begin{table}[]
    \centering
    \resizebox{0.6\columnwidth}{!}{%
    \begin{tabular}{@{}ccccc@{}}
    \toprule
    \textbf{Dataset} &
    \textbf{Architecture} &
    \textbf{ELBO} &
    \textbf{Rate} &
    \textbf{Distortion} \\
     \midrule
    \multirow{4}{*}{MNIST} & 
     ResNet &  89.63 & 29.17 & 60.46 \\
     & CNN & 96.35 & 18.33 & 78.03  \\
     & MR-ResNet & \textbf{87.89} & 27.07 & 60.82  \\
     & MR-CNN & 93.51 & 28.50 & 65.00  \\
     \midrule
    \multirow{4}{*}{Omniglot} & 
     ResNet & 111.07 & 34.98 & 76.09 \\
     & CNN & 114.73 & 34.11 & 80.62  \\
     & MR-ResNet & \textbf{107.78} & 34.36 & 73.41  \\
     & MR-CNN & 110.98 & 34.21 & 76.77  \\
     \midrule
    \end{tabular}
    }
    \vspace{0.1cm}
    \caption{We present the lowest test ELBO scores for $\beta$-VAE and MR-VAE on MNIST and Omniglot datasets.}
    \vspace{0.2cm}
    \label{tab:elbo_score}
\end{table}

\subsection{Image Reconstruction}
\label{appsub:image_reconstruction}

We used binary static MNIST~\citep{larochelle2011neural}, Omniglot~\citep{lake2015human}, CIFAR-10~\citep{krizhevsky2009learning}, SVHN~\citep{netzer2011reading}, and CelebA64~\citep{liu2015faceattributes} datasets. 

We used the Adam optimizer for CNN and ResNet~\citep{he2016deep} architectures for 200 epochs with a batch size of 128 and a cosine learning rate decay. For NVAE architecture with binary images, we used the Adamax optimizer for 400 epochs with a batch size of 256 and a cosine learning rate decay. In the case of CelebA64 dataset, we use a batch size of 16. For all baseline models, we conducted hyperparameter searches over learning rates \{0.01, 0.003, 0.001, 0.0003, 0.0001, 0.00003, 0.00001\} with $\beta = 1$, making choices based on the final validation loss. With the chosen set of hyperparameters, we repeated the experiments over 10 $\beta$s uniformly sampled between 0.01 and 10 on a log scale with and without the KL warm-up. The experiments were repeated 3 times with different random seeds and the mean is presented.

We followed the same architectural configurations from~\citet{chadebec2022pythae} for CNN and ResNet architectures\footnote{\label{footnote:pythae}\url{https://github.com/clementchadebec/benchmark_VAE}}. We also adopted the NVAE architecture and configurations from~\citet{vahdat2020NVAE}\footnote{\url{https://github.com/NVlabs/NVAE}}.

MR-VAEs were trained using the same training and architectural configurations as the baseline. We performed a grid search on learning rates \{0.01, 0.003, 0.001, 0.0003, 0.0001, 0.00003, 0.00001\} and selected the learning rate that achieved the best validation loss at $\beta = 1$. With different random seeds, the experiment was repeated 3 times and the mean was shown.

Note that, for NVAE architecture on the CelebA64 dataset, we observed instability in the initial stage of training. To stabilize the early training stage, we used a warm-up stage for the initial 30\% of the training where the KL weight remained fixed to a small constant.

In addition, we used the Yahoo (Answer)~\citep{zhang2015character} dataset and adopted the training procedure from~\citet{hu2019texar}. We trained the network for 200 epochs with a batch size of 32 using the Adam optimizer. The learning rate was decayed by a factor of 2 when the validation loss did not improve for 2 epochs. For all baseline methods, we performed hyperparameter searches over the learning rates \{0.01, 0.003, 0.001, 0.0003, 0.0001, 0.00003, 0.00001\} with the KL warm-up and $\beta = 1$ and selected configuration that achieved the lowest validation loss. With the chosen set of hyperparameters, we repeated the experiments over 10 $\beta$s uniformly sampled between 0.01 and 10 on a log scale with and without the KL warm-up.

We also adopted the architectural configurations from~\citet{hu2019texar}\footnote{\url{https://github.com/asyml/texar-pytorch}}. LSTM models used the embedding dimension of 256 and the hidden dimension of 256. In training MR-VAEs, we followed the same configurations as in the baseline models. We performed a grid search on learning rates in the range \{0.01, 0.003, 0.001, 0.0003, 0.0001, 0.00003, 0.00001\} and selected the learning rate that achieved the best validation loss at $\beta = 1$. We report the mean across 3 different random seeds.

\subsubsection{$\beta$-TCVAE}

We used the dSprites~\citep{dsprites17} dataset and followed the same training and architectural configurations from~\citet{chen2018isolating}\footnote{\url{https://github.com/rtqichen/beta-tcvae}}. We trained the network for 50 epochs with the Adam optimizer and a batch size of 2048. For the baseline method, we performed grid search over the learning rates \{0.01, 0.003, 0.001, 0.0003, 0.0001, 0.00003, 0.00001\} with $\beta = 1$ and selected configuration that achieved the lowest training loss. With the chosen training configuration, we repeated the experiments over 10 $\beta$s uniformly sampled between 0.1 and 10 on a log scale. We report the mean across 3 different random seeds.

The encoder is composed of 2 fully-connected layers with a hidden dimension of 1200 and ReLU activation functions. Similarly, the decoder consists of two fully-connected layers, but the Tanh activation function is used. A latent dimension of 10 is used. We followed the same grid search for training MR-VAEs and selected the learning rate that achieved the best training loss at $\beta = 1$. We also report the mean across 3 different random seeds.

\begin{figure}[!t]
    \centering
    \begin{tabular}{ccc}
    \includegraphics[width=0.32\linewidth]{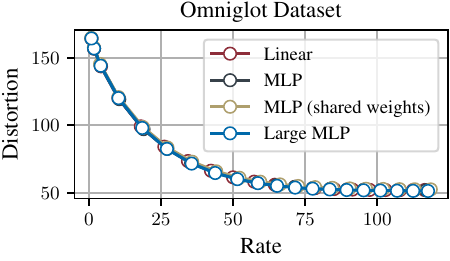}
    \includegraphics[width=0.32\linewidth]{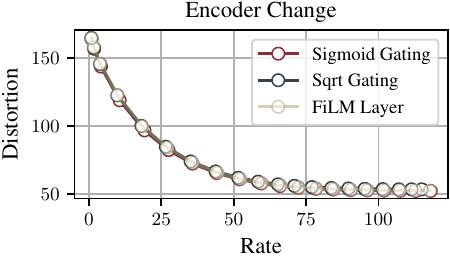}
    \includegraphics[width=0.32\linewidth]{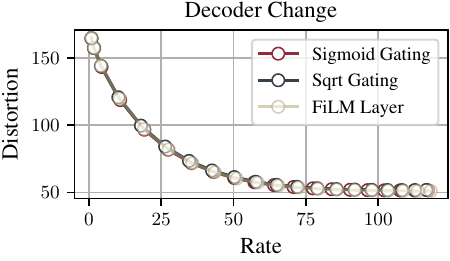}
    \end{tabular}
    \caption{\small \textbf{(left)} An ablation studying the effect of different architectural designs for the hypernetwork. \textbf{(middle and right)} An ablation studying the effect of using different gating mechanisms on the hypernetwork activation. The MR-VAEs were trained on the Omniglot dataset.}
    \label{fig:ablation}
\end{figure}

\section{Additional Experiments}
\label{app:additional_experiments}

\subsection{Ablation Study}

We further investigate the effect of modifying the architecture and activations of MR-VAE architectures. We used the ResNet encoder and decoder trained on the Omniglot dataset, following the same setup from the image reconstruction experiment. 

\paragraph{Ablations over Architecture Design.} In MR-VAEs, we apply an affine transformation to the $\log \beta$ to scale each layer's pre-activations. Here, we investigate if more expressive architectural designs can help in learning better rate-distortion curves. We repeated the experiment by changing the architecture for the scaling function, where we used linear transformation (default), a 2-layer MLP (with and without weight sharing on the first layer), and a 5-layer MLP. As shown in Figure~\ref{fig:ablation}, we observed that increasing the capacity of the model does not improve the final performance on the rate-distortion curve. 

\paragraph{Ablations over Activation Function.} Next, we investigated the effect of different design choices for gating the pre-activations. We repeated the experiment with sigmoid gating (default for encoders), sqrt gating (default for decoders), and FiLM layer~\citep{perez2018film} (which is similar to YOTO \citep{dosovitskiy2019you}). As shown in Figure~\ref{fig:ablation}, while MR-VAE's default choice performs slightly better compared to other combinations, we observed that the choice of the gating mechanism does not significantly impact the final performance on rate-distortion curves.

\end{document}